\documentclass[preprint,review,1p,times,sort&compress]{elsarticle}

\usepackage{amssymb}
\usepackage{amsmath}
\usepackage{pifont}

\usepackage{algorithm}
\usepackage{algorithmicx}
\usepackage{algcompatible}

\usepackage{xcolor} 
\usepackage{booktabs}
\usepackage{multirow}
\usepackage{hyperref}

\journal{Pattern Recognition}

\begin{document}

\begin{frontmatter}



\title{MoTiC: Momentum Tightness and Contrast for Few-Shot Class-Incremental Learning} 



\author{Zeyu He\textsuperscript{a,1},
        Shuai Huang\textsuperscript{b,1},
        Yuwu Lu\textsuperscript{b,*},
        Ming Zhao\textsuperscript{c}
        }



\fntext[label2]{
    These authors contributed equally to this work.
}

\cortext[cor1]{
    School of Artificial Intelligence, South China Normal University, Foshan, 528225, China. \\
    \hspace*{1.5em} \textit{E-mail} address: \href{mailto:luyuwu2008@163.com}{luyuwu2008@163.com} (Y. Lu).
    }

\affiliation[label1]{organization={School of Computer Science and Information Security},
            addressline={Guilin University of Electronic Technology},
            city={Guilin},
            country={China}}
\affiliation[scnu]{organization={School of Artificial Intelligence},
addressline={South China Normal University}, 
city={Foshan},
country={China}}

\affiliation[wuxi]{organization={School of Internet of Things Engineering},
addressline={Wuxi University}, 
city={Wuxi},
country={China}}



\begin{abstract}
Few-Shot Class-Incremental Learning (FSCIL) must contend with
the dual challenge of learning new classes from scarce samples
while preserving old class knowledge.
Existing methods use the frozen feature extractor and class-averaged prototypes to mitigate against catastrophic forgetting and overfitting.
However, new-class prototypes suffer significant estimation bias due to extreme data scarcity, whereas base-class prototypes benefit from sufficient data.
In this work, we theoretically demonstrate that aligning the new-class priors with old-class statistics via Bayesian analysis reduces variance and improves prototype accuracy.
Furthermore, we propose large-scale contrastive learning to enforce cross-category feature tightness.
To further enrich feature diversity and inject prior information for new-class prototypes, we integrate momentum self-supervision and virtual categories into the Momentum Tightness and Contrast framework (MoTiC), 
constructing a feature space with rich representations and enhanced interclass cohesion.
Experiments on three FSCIL benchmarks produce state-of-the-art performances,
particularly on the fine-grained task CUB-200, validating our method's ability to reduce estimation bias and improve incremental learning robustness.
Our code is available at \url{https://github.com/huangshuai0605/MoTiC}.
\end{abstract}




\end{frontmatter}



\section{Introduction}
\label{Introduction}
Humans exhibit a high level of transfer learning capability, enabling them to generalize knowledge from previous experiences to novel situations, which is called cognitive learning \cite{NatCommun2025}. This ability allows humans to quickly grasp new categories from a small number of examples. However, machine learning mechanisms still do not align well with the principles of cognitive learning \cite{NatCommun2025}.

In real-world scenarios, data distributions and categories continually evolve, however, acquiring large amounts of annotated data remains highly challenging. A new machine learning mechanism is needed to assist models in adapting to these variations, enabling swift adjustments and optimization of their performance. To address this challenge, Few-Shot Class-Incremental Learning (FSCIL) \cite{FewShotCIL,SurveyFSCIL,FILP3D} is proposed to design artificial intelligence systems that can learn new classes from only a few samples, while preserving performances on previously encountered classes.

However, the FSCIL scenario poses two significant challenges: catastrophic forgetting (loss of previously acquired knowledge) and overfitting (overfitting on few training samples and thus producing poor generalizability). To address these convoluted issues and facilitate the model's flexible adaptation, most previous studies \cite{CLOM,EvolvedClassifiers,OpenSetFSCIL,NeuralCollapseFSCIL,SAVC} freeze the backbone after training on the base classes and employ evolving classifiers, utilizing feature-averaged class prototypes as embeddings of the classifiers. However, this formation strongly depends on the representations obtained via softmax cross-entropy loss on the base classes, which often causes neural collapse \cite{PNAS2020} and a loss of generalizability of the unseen classes. In the realm of representation learning, particularly in self-supervised contrastive learning (SSC), significant breakthroughs have been achieved in enhancing the transferability of learned representations to downstream tasks. 

In this work, we introduce the Momentum Tightness and Contrast framework (MoTiC), a novel methodology founded on several essential foundations. These foundations include moment self-supervised contrastive learning and the mutual tightness of features on the representation hypersphere facilitated by a large, consistent feature queue, and also the enrichment at the class level through virtual categories.

Based on our Bayesian analysis, compactness among categories facilitates the reuse of features between them, increasing the accuracy of new-class prototypes under few-shot sampling, thereby leading to the emergence of new classes. For example, the representation of an owl can be a combination of cat-like eyes and the body of an eagle. This approach is particularly suitable for fine-grained datasets \cite{PanPFR}, where new classes are often composed of base-class features. Furthermore, maintaining a large and consistent dictionary can encompass the entire feature space of the encoder as much as possible, allowing greater integration of features across different categories, and contributing to the robustness of the encoder's representations. Our research contributions are summarized as follows:

\begin{figure}[t]
\centering
\includegraphics[width=0.8\textwidth]{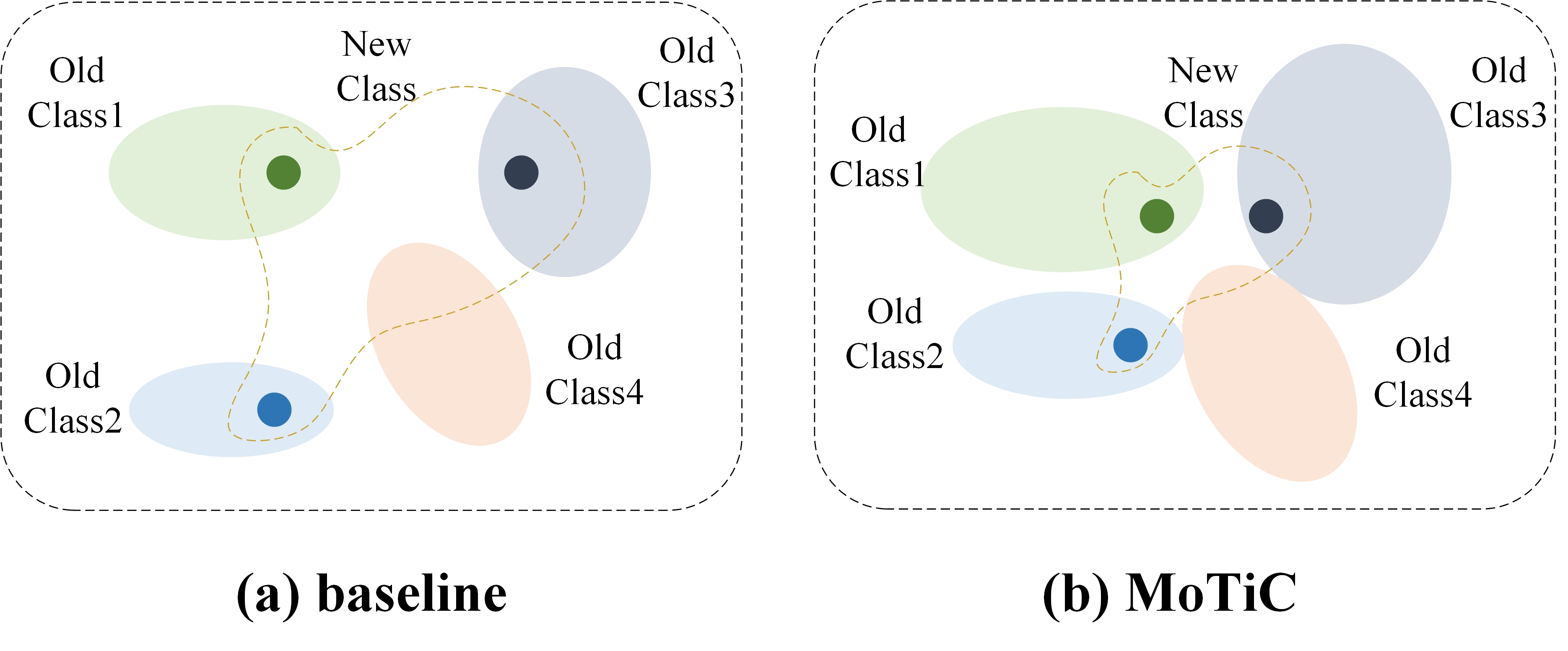}
\caption{The MoTiC framework enhances the Bayesian prior information of old class prototypes for new classes, significantly increasing the accuracy of new class prototypes under few-shot sampling scenarios.}\label{first}
\end{figure}

\begin{itemize}
    \item We propose a momentum tightness and contrast framework (MoTiC) that provides an excellent solution for FSCIL through the variety and tightness of the feature space.
    \item Our Bayesian analysis demonstrated that controlling interclass distances enhances the generalizability to novel classes.
    \item We significantly improve the performance of FSCIL on three commonly used benchmark tests \cite{CUB2011,CIFAR100,miniImageNet} and achieve state-of-the-art results on both the CUB200 and CIFAR100 datasets.
\end{itemize}

The structure of the paper is as follows. 
Section \ref{Related works} introduces related research on Few-Shot Class-Incremental Learning, Self-supervised learning, and Transferable representation learning. 
Section \ref{Background} presents the problem formulation, essential groundwork, and related analysis. 
Section \ref{Method} presents our proposed approach in detail. 
Section \ref{Experiments} outlines the experimental setup and presents the results validating the effectiveness of our proposed method. 
Section \ref{Conclusion} summarizes this paper.

\section{Related works}
\label{Related works}

\subsection{Few-shot class-incremental learning}
\label{Few-shot class-incremental learning}

The FSCIL paradigm aims at continually learning from limited labelled samples. The methods in FSCIL can be categorized into replay or distillation strategies
\cite{LCwoF,ClassAwareDistillation}, metric learning approaches \cite{SAVC,FACT,CompositionalFSCIL}, meta-learning methods \cite{NeuralCollapseFSCIL,MetaFSCILApproach}, etc, highlighting the diversity of FSCIL methodologies. In parallel with our approach, metric learning approaches demonstrate various ways to adapt to unseen classes by learning the feature distribution of the data. SAVC \cite{SAVC} introduces virtual classes into supervised contrastive learning (SCL) to act as placeholders for unseen classes. FACT \cite{FACT} assigns virtual prototypes to compress the embeddings of seen classes and reserves space for new classes. Comp-FSCIL \cite{CompositionalFSCIL} constructs a compositional model to enhance the reusability of base class features. Most of the studies freeze the feature extractor trained on the base session and employ an evolving classifier using class prototypes to significantly address catastrophic forgetting and overfitting issues, and 
they all encourage greater separation among the base classes. In contrast to encouraging class differentiation, we propose enhancing the richness of intraclass features and promoting tightness between class features to generate new classes, which improves the recognition accuracy for unseen categories. Similarly, Zou \cite{CLOM} and Kim \cite{CLOSER} emphasize the effectiveness of reducing interclass distances or negative boundary softmax loss for FSCIL. 

In comparison, our method more effectively considers the global feature distribution by maintaining intraclass richness and interclass compactness on a large, consistent feature queue, introducing virtual class as placeholders to provide new classes with prior information.

\subsection{Self-supervised learning}
\label{Self-supervised learning}

Unsupervised visual representation learning \cite{NPID,MoCo,SimCLR,Bootstrap,SimSiam,DeepCluster,SWAV,SPARE} with contrastive loss has made significant advancements, even surpassing supervised counterparts in many downstream tasks. Instance discrimination tasks and clustering-based tasks are two typical pretext tasks in this research domain.
Instance discrimination \cite{NPID,MoCo,SimCLR,Bootstrap,SimSiam} tasks mostly differ in maintaining negatives of varying sizes. In the initial method \cite{NPID}, a large amount of memory is employed to store the features of all the images in the dataset. However, the features lack consistency during training. He et al. \cite{MoCo} introduce a momentum encoder with a queue to maintain a large and consistent set of negative sample features. Chen et al. \cite{SimCLR} implement end-to-end contrastive learning via large batch training supported by TPUs, incorporating a projection layer. Grill et al. \cite{Bootstrap} propose a method for contrastive learning without any negatives. Chen et al. \cite{SimSiam} highlight the crucial role of the stop-gradient operation in preventing representation collapse during contrastive learning.
The objective of clustering-based tasks is to learn visual representations and cluster them together. DeepCluster \cite{DeepCluster} is a typical approach that uses convolutional networks to extract features for clustering and label assignment. Carnon et al. \cite{SWAV} introduced an online algorithm, SwAV, which simultaneously clusters the data while performing contrastive learning in small batches.

Research has advanced the progress of the SSC approaches. In this study, we aim to explore the application of several mainstream SSC methods, such as MoCo, in the context of FSCIL.

\subsection{Transferable representation learning}
\label{Transferable representation learning}
The quest for high-quality feature representations that can transfer across various downstream tasks has garnered significant attention for an extended period \cite{BEI2025,ImageNet Transfer, TransferFeatures,contrastTransfer,NegativeMargin}. Kornblith et al. \cite{ImageNet Transfer, TransferFeatures} report that models or loss functions that achieve the best classification performance on the ImageNet dataset often exhibit low transferability to other downstream tasks. Several methods \cite{MoCo, SimCLR, Bootstrap,SimSiam, SWAV, SPARE} have reported that the representations obtained through self-supervised contrastive learning (SSC) demonstrate strong transferability. Moreover, Ericsson et al. \cite{SelfSupervisedTransfer} conduct a comprehensive comparison of various SSC methods regarding their transferability. Islam et al. \cite{contrastTransfer} discovered that SSC contrastive methods encapsulate the richest low-level and mid-level information and enhance intraclass variability, which is beneficial for transferring to other tasks. Liu et al. \cite{NegativeMargin} propose the incorporation of negative margins into the softmax function to aid in recognizing new classes with few samples.

Building on prior insights, we integrate self-supervised learning with large-batch contrastive learning to compress interclass margins, which are augmented by virtual category placeholders for enhanced representation transferability in FSCIL scenarios.

\section{Background}
\label{Background}

\subsection{Problem definition}
\label{Problem definition}
FSCIL trains models on a base session that contains many labelled training samples, continuously extending the model to incremental sessions with only a few labelled training samples. There is a constant stream of training data indicating $\mathcal{D}_{\text{train}} = \{\mathcal{D}_{\text{train}}^t\}_{t=0}^T$, where $ \mathcal{D}_{\text{train}}^t = \{(x_i, y_i)\}_{i=0}^{N_t} $ denotes the training samples from session $t$, and where $x_{i}$ and $y_{i}$ are the $i$-th image and corresponding label, respectively. Supposing that the label space for the $t^{th}$ session is $C^{t}$, the label space from different sessions has no intersection. In each $t^{th}$ training session, only the corresponding dataset $D^{t}$ is accessible for training. The model trained on $\mathcal{D}_{train}^{t}$ should be evaluated on $\mathcal{D}_{test}^{t}$, which contains all the classes encountered in the $t^{th}$ session, specifically $C^0 \cup C^1 \cup \ldots \cup C^t$. In particular, the incremental training data are organized in an $N$-way $K$-shot format, which consists of $N$ classes, each containing $K$ training images.

\subsection{Baseline}
\label{Baseline}
Several significant works \cite{CLOM,EvolvedClassifiers,SAVC,CLOSER,LIU2025} recommend an incremental-frozen framework that initially leverages sufficient data in the base session to train the model by optimizing each sample loss as:

\begin{equation}
\mathcal{L}_{ce}(f_\theta,w; B,\tau) =
-\frac{1}{B} \sum_{i=1}^{B}
\log\left(\frac{\exp\left(\text{sim}(f_{\theta}(x_i),w_{y_i})/\tau\right)}{\sum_{k=1}^{C^0}\exp\left(\text{sim}(f_{\theta}(x_i),w_k)/\tau\right)}\right),
\end{equation}
where $B$ is the batch size, $\mathcal{L}_{ce}$ is the cross-entropy loss (CE), $f_{\theta}$ represents the feature extractor with parameters $\theta$, $w_{i}$ denotes the class prototypes of class $i$, and sim is the cosine similarity function between two vectors: $\text{sim}(x,y) = {x^Ty}/(\|x\|\|y\|)$ and $\tau$ is a hyper temperature parameter. 

In the subsequent increment phase, the feature extractor $f_{\theta}(x)$ is completely frozen, and the classifier is expanded in each incremental learning session:
\begin{equation}
    W = \{w_1^{(0)}, w_2^{(0)}, \ldots, w_{|C^0|}^{(0)}\}, \ldots, \{w_1^{(t)}, w_2^{(t)}, \ldots, w_{|C^t|-|C^{t-1}|}^{(t)}\}.
\end{equation}
The weights of the classifier are attributed to the average embeddings of each class ($i.e.$, the prototype), a process that we refer to as categorical embedding averaging (CEA), represented as:
\begin{equation}
w_c^t = \frac{1}{n_c^t} \sum_{i=1}^{n_c^t} {f(x_{c,i})},
\end{equation}
where $n_{c}^{t}$ is the number of training samples in connection with the $c^{th}$ class and session $t$. $x_{c,i}$ is the $i^{th}$ training sample of the $c^{th}$ class.

In each session, the inference process is executed via the nearest class mean (NCM) algorithm \cite{DistanceBasedGen} to assess the accuracy across all encountered classes. Specifically, the algorithm calculates the proximity between the feature $f_{\theta}(x)$ and all the prototypes in the classifier. This proximity is quantified by the cosine similarity, which is defined as:
\begin{equation}
    c_x = \arg\max_{c,t} \text{sim} \left( f_{\theta}(x), w_c^t \right).
\end{equation}

\subsection{Bayesian estimation for baseline}
\label{Bayesian estimation for baseline}
The incremental-frozen baseline faces an obvious problem: prototypes of the base classes, which are sampled from a large amount of data, are accurate, whereas prototypes of the new classes suffer from substantial estimation bias due to extreme data scarcity. 
Using Bayesian parameter estimation, we formalize how semantic similarity between classes acts as a prior to refine new-class prototypes under data scarcity.

Let $\mathcal{D}_c = \{f_{\theta}(x_1), f_{\theta}(x_2), \ldots, f_{\theta}(x_K) \} $ denote $K$ support samples for a new class $c$, where $f_{\theta}(.)$ is the feature extractor. Without prior knowledge, the prototype estimate is as follows:
\begin{equation}
\hat{\mu}_c^{\text{MLE}} = \frac{1}{K} \sum_{i=1}^{K} f_{\theta}(x_i),
\end{equation}
which suffers from high variance $\frac{\sigma^2}{K}$ when $K$ is small, where $\sigma^2$ denotes the inherent noise variance of
the feature encoder $f_{\theta}$ output, $i.e.$, $f_{\theta}(x_i) \sim \mathcal{N}(\mu_c, \sigma^2 I)$.
To mitigate this, we introduce a similar old class $c^{\prime}$
with a well-estimated prototype. Assuming a Gaussian prior
$\mu_c \sim \mathcal{N}(\mu_{c'}, \tau^2 I)$, where $\tau^2$ quantifies the prior knowledge from $c^{\prime}$ ($e.g., \tau^2 $ decreases as classes become closer or as the features of class $c^{\prime}$
becomes richer), the Bayesian posterior estimate becomes:
\begin{equation}
    \hat{\mu}_c^{\text{Bayes}} = \frac{\frac{1}{\sigma^2} \sum_{i=1}^{K} f_{\theta}(x_i) + \frac{1}{\tau^2} \mu_{c'}}{\frac{K}{\sigma^2} + \frac{1}{\tau^2}},
\end{equation}
where $\sigma^2$ is the sample variance. The posterior variance is reduced to:
\begin{equation}
    \text{Var}(\hat{\mu}_c^{\text{Bayes}}) = \left( \frac{K}{\sigma^2} + \frac{1}{\tau^2} \right)^{-1},
\end{equation}
which is significantly lower than the MLE variance $\frac{\sigma^2}{K}$ when $\tau^2$ is smaller, as shown in in Fig. \ref{first}. This finding mathematically explains why more prior information enhances estimation accuracy under limited samples.

\section{Method}
\label{Method}
\begin{figure}[t]
\centering
\includegraphics[width=1.0\textwidth]{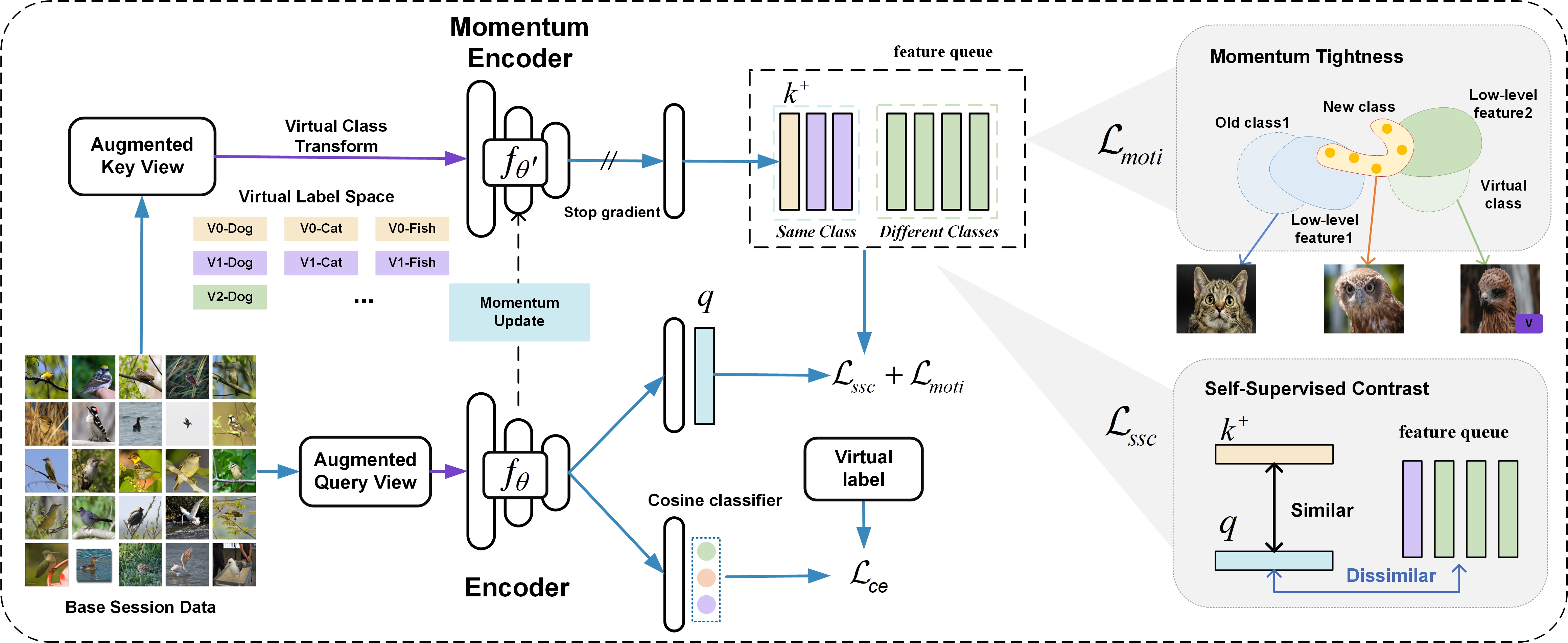}
\caption{MoTiC Model Architecture. The MoTiC framework maintains a large and consistent feature dictionary and introduces virtual classes, enhancing feature richness and tightness across multiple levels of fine granularity.}
\label{model}
\end{figure}

The Bayesian analysis of the baseline demonstrates that enhancing the accuracy of few-shot prototypes can be achieved by providing richer prior information. Inspired by this insight, we propose enhancing feature richness, reducing interclass distances, and introducing virtual categories, 
thereby offering prior information for novel class prototypes from three complementary aspects.  Fig. \ref{model} shows the learning process of the proposed method.

\subsection{Improving feature richness}
\label{Improving feature richness}
We adopt the strategy from MoCo \cite{MoCo} to improve feature richness in our method. This strategy allows for the storage of slowly evolving features from recent batches within a large and consistent feature dictionary while maintaining a small budget. For each image-label pair $(x,y)$, we obtain the query view $x_q = Aug_q(x)$ and the key view $x_k = Aug_k(x)$ through data augmentation. The two views are subsequently input into the query encoder $f_{\theta}$ and the key encoder $f_{\theta^{\prime}}$, producing two representations $q$ and $k^{+}$ after $L_2$ normalization. Notably, $f_{\theta}$ and $f_{\theta^{\prime}} $ consist solely of a feature extractor without a projector, which we observe, will lead to performance degradation. Both encoders share the same architecture, with the key encoder not participating in backpropagation and instead updating its parameters by aggregating them with a large momentum from the query encoder: $\theta' \leftarrow m\theta' + (1 - m)\theta.$ We maintain a feature queue $Q$ and a label queue of the same length to store the key features and its corresponding labels from the most recent batch. Consequently, we obtain the following self-supervised contrastive loss for a batch:
\begin{equation}
    L_{ssc}(f_\theta; B, \tau_{v}, A) = -\frac{1}{B} \sum_{i=1}^{B} \log \frac{\exp(q_i^T k_i^+ / \tau_{v})}{\sum_{k' \in A(k_i^+)} \exp(q_i^T k' / \tau_{v})}, 
\end{equation}
where $A(.) =. \cup Q$ is the concatenation of the key features $k^{+}$ of $x$ and feature queue $Q$, $\tau$ is a temperature hyper-parameter coefficient, and $B$ is the batch size.

\begin{algorithm}[t]
    \caption{Momentum Tightness and Contrast}
    \label{alg:motic}
    $\mathbf{Input:}$ query encoder $f_q$, key encoder $f_k$, feature queue $f_{queue}$ ($\mathbb{R}^{C \times K}$), label queue $l_{queue}$ ($\mathbb{R}^{K}$), momentum $m$ and other related hyperparameters
    
    $\mathbf{Output:}$ trained query encoder $f_q$
    \begin{algorithmic}[1]
    \STATE Initialize $f_k$ with params of $f_q$ 
    \WHILE{stop criteria is not satisfied}
        \STATE \text{Obtain a batch } $\{x_i, y_i\}_{i=1}^n$ from $\mathcal{D}$
        \STATE Generate transformed virtual samples $x_{\mathcal{T}},y_{\mathcal{T}}$: $\{x_{i}^{m},y_{i}^{m}\}_{m=1}^M$ by virtual transformations set $\mathcal{T}$
        \STATE Compute: $q_{\mathcal{T}} \gets f_q(Aug_q(x_{\mathcal{T}}))$ and $k_{\mathcal{T}} \gets f_k(Aug_k(x_{\mathcal{T}}))$
        \STATE Calculate cross-entropy loss $\mathcal{L}_{ce}(x_{\mathcal{T}},y_{\mathcal{T}} )$ by Eq. (\ref{eq:ce_loss})
        \STATE Calculate metric learning losses $\mathcal{L}_{ssc}(q_{\mathcal{T}},k_{\mathcal{T}})$  by Eq. (\ref{eq:ssc_loss})
        \STATE Calculate metric learning losses $\mathcal{L}_{moti}(q_{\mathcal{T}},k_{\mathcal{T}})$ by Eq. (\ref{eq:moti_loss})
        \STATE Minimize loss: $\mathcal{L} = \mathcal{L}_{ce} + \lambda_{ssc}\mathcal{L}_{ssc} + \lambda_{moti}\mathcal{L}_{moti} $ in Eq. (\ref{eq:total_loss})
        \STATE Update $f_q$ from the backward of $\mathcal{L}$ 
        \STATE Update $f_k$ with momentum $m$ using $f_q$ parameters
        \STATE Update $f_{queue}$ with $k_{\mathcal{T}}$ by enqueue-dequeue operation
        \STATE Update $l_{queue}$ with $y_{\mathcal{T}}$ by enqueue-dequeue operation
    \ENDWHILE
    \end{algorithmic}
\end{algorithm}

\subsection{Narrowing interclass gap}
\label{Narrowing inter-class gap}
Although the additional content contained in new classes is invisible in the base session, the model in the FSCIL should be self-sufficient;  it should inherently provide sufficient prior knowledge for novel class prototypes, whereas reducing interclass gaps can offer adequate prior information for new classes. Novel classes emerge at the intersections of distinct base-class feature distributions, synthesizing cross-category characteristics. In other words, the model should aggregate its learned low-level or mid-level discriminative features (e.g., cat eyes and avian morphologies) to compositionally construct novel class representations (e.g., owls), thereby achieving semantically grounded prototype generation through feature space recombination.

In the MoTiC framework, we employ a different approach from the conventional supervised contrastive methods, which typically aim to enhance the cohesion of the same class in the feature space. Instead, we encourage the tightness of different classes to better combine the rich subfeatures obtained from SSC, as well as to provide more prior information for the prototypes of new classes. Specifically, we introduce a novel supervised contrastive loss function $\mathcal{L}_{MoTi}$ to promote the aggregation of category features. For a query encoded as $q$ with category $c_{q}$, we desire it to be more adjacent to features in the feature queue Q, which belong to different classes:
\begin{equation}
\mathcal{L}_{MoTi}(f_\theta; B, F) = -\frac{1}{B} \sum_{i=1}^{B} \frac{\sum_{k' \in F(q_i, k^+)} q_i^T k'}{|F(q_i, k^+)|},
\end{equation}
where $F(q, k^+) = \{ k' \in (k^+ \cup Q) : c(k') \neq c(q) \}$ represents the features that are different in category from $q_{i}$, which are obtained by concatenating $k^{+}$ from the batch with the feature queue $Q$.

\subsection{Introducing virtual classes}
\label{Introducing virtual classes}
Bayesian analysis of the baseline indicates that if added virtual classes share semantic similarities with novel classes, the quality of novel class prototype estimations can be considerably enhanced through prior information. This mechanism is mainly achieved by transferring prior knowledge driven by interclass similarity. Furthermore, the introduction of these virtual categories can synergize with self-supervised and momentum tightness mechanisms, collectively achieving superior performances.

In the MoTiC framework, we draw inspiration from \cite{SAVC} to generate virtual classes by applying predefined image transformations ($i.e.,$ rotations) within supervised contrastive learning, assuming that there are $M$ transformations in the transformation set $\mathcal{T}$. We then generate $M$ transformed samples for each image-label pair $(x,y)$, denoted as $\left\{ \left( x_m, y_m \right) \right\}_{m=1}^{M}$, where $y_m = y \times M + m$ and the subscript of $( x_m, y_m)$ corresponds to the $m$-th transformation applied to $(x,y)$. In this way, we expand the label space by a factor of $M$, allowing each virtual class to occupy new positions in the feature space, and providing novel semantic information for the prototypes of new classes. We conduct a more fine-grained training process during classification, self-supervised learning, and our momentum tightness mechanism:
\begin{equation}
\label{eq:ce_loss}
\mathcal{L}_{ce}(f_\theta,w;B,\tau,\mathcal{T}) = -\frac{1}{B M} \sum_{i=1}^B \sum_{m=1}^M \log \left( \frac{\exp\left( \operatorname{sim}(f_\theta(x_{i,m}), w_{y_{i,m}}) / \tau \right)}{\sum_{k=1}^{C^0} \sum_{n=1}^M \exp\left( \operatorname{sim}(f_\theta(x_{i,m}), w_{(k,n)}) / \tau \right)} \right),
\end{equation}

\begin{equation}
\label{eq:ssc_loss}
\mathcal{L}_{ssc}(f_\theta; B, \tau_{v},A,\mathcal{T}) = - \frac{1}{B M} \sum_{i=1}^{B} \sum_{m=1}^{M} \log \frac{
    \exp\left( q_{i, m}^{T} k_{i, m}^{+} \, / \, \tau_{v} \right)
}{
    \sum_{k' \in A\left( k_{i, m}^{+} \right)} 
    \exp\left( q_{i, m}^{T} k' \, / \, \tau_{v} \right)
},
\end{equation}

\begin{equation}
\label{eq:moti_loss}
\mathcal{L}_{MoTi}(f_\theta; B,F,\mathcal{T}) = 
-\frac{1}{B M} 
\sum_{i=1}^{B} \sum_{m=1}^{M} 
\frac{
    \sum\limits_{k' \in F\left( q_{i,m}, k^{+} \right)} 
    q_{i,m}^{T} k'
}{
    \left| F\left( q_{i,m}, k^{+} \right) \right|
},
\end{equation}
where $x_{i,m}=\mathcal{T}_{m}(x_i)$ with query feature $q_{i,m} = f_{\theta}(Aug_{q}(x_{i,m}))$ and key feature $k_{i,m}^{+} = f_{\theta^{\prime}}(Aug_k(x_{i,m}))$. The overall training objective can be written as:
\begin{equation}
    \label{eq:total_loss}
    \mathcal{L} = \mathcal{L}_{ce} + \lambda_{ssc} \mathcal{L}_{ssc} + \lambda_{MoTi} \mathcal{L}_{MoTi}.
\end{equation}

\subsection{Multigrained fusion inference}
\label{Multi-grained fusion inference}
In traditional classification tasks, models typically rely on a single prototype per class for inference. However, as the number of classes increases, fine-grained semantic information (e.g., different perspectives of objects) becomes critical for classification performance. For example, when distinguishing between two similar birds, the model must capture fine-grained features such as the frontal, lateral, and dorsal views. To address this, we propose leveraging the multigrained information obtained during the MoTiC training phase to aggregate inference results from diverse fine-grained semantics, thereby enhancing classification accuracy.

First, we extend the prototype library by generating $M$ fine-grained prototypes for each class through transformation $\mathcal{T}$: $W = \bigcup_{t=0}^{T} \bigcup_{c=1}^{|C^t|} \{ w_{c1}^t, w_{c2}^t, \ldots,$ $w_{cM}^t \},$ where $w_{cm}^t$ denotes the $m$-th fine-grained prototype of class $c$ at the t-th session. Then, prototypes sharing the same fine-grained semantic index $m$ form a conditional subset $W_{m}$, representing clustering centres for a specific semantic granularity ($e.g.$, frontal view): $W_m = \bigcup_{t=0}^{T} \bigcup_{c=1}^{|C^t|} \left\{ w_{cm}^t \right\}.$ Finally,
for a test sample $x$, we generate $M$ transformed features $\mathcal{T}(x) = \{x_1,x_2,\ldots\,x_M\}$, each aligned with a specific semantic subset $W_m$. The final classification is derived by aggregating the similarities across all the subsets:
\begin{equation}
    c_x = \arg\max_{c,t} \sum_{m=1}^{M} \text{sim} \left( f(x_m), w_{cm}^t \right).
\end{equation}
As anticipated, the virtual classes and multigrained fusion predictions effectively provide sufficient prior information for novel classes, achieving superior predictive performances for the new classes.

\begin{table*}[t]
    \caption{10-way 5-shot performance on CUB200 using the pretrained Resnet18.}
    \label{tab:cub200_results}
    \centering
    \resizebox{1.0\textwidth}{!}{%
    \begin{tabular}{lcccccccccccr}
    \hline
    \multicolumn{1}{c}{\multirow{2}{*}{Method}} & \multicolumn{11}{c}{Acc. in each session (\%)} & \multicolumn{1}{c}{\multirow{2}{*}{Avg (\%)}} \\ \cline{2-12}
    \multicolumn{1}{c}{} & 0 & 1 & 2 & 3 & 4 & 5 & 6 & 7 & 8 & 9 & 10 & \multicolumn{1}{c}{} \\ \hline
    Baseline & 79.92 & 76.23 & 73.18 & 69.45 & 67.83 & 65.74 & 64.54 & 63.33 & 61.56 & 61.27 & 60.10 & 67.55 \\ \hline
    TOPIC \cite{FewShotCIL} & 68.68 & 62.49 & 54.81 & 49.99 & 45.25 & 41.40 & 38.35 & 35.36 & 32.22 & 28.31 & 26.26 & 43.92 \\
    F2M \cite{FlatMinima} & 81.07 & \underline{78.16} & \underline{75.57} & \underline{72.89} & \underline{70.86} & \underline{68.17} & \underline{67.01} & 65.26 & 63.36 & 61.76 & 60.26 & \underline{69.48} \\
    CEC \cite{EvolvedClassifiers} & 75.85 & 71.94 & 68.50 & 63.50 & 62.43 & 58.27 & 57.73 & 55.81 & 54.83 & 53.52 & 52.28 & 61.33 \\
    ALICE \cite{OpenSetFSCIL} & 77.40 & 72.70 & 70.60 & 67.20 & 65.90 & 63.40 & 62.90 & 61.90 & 60.50 & 60.60 & 60.10 & 65.74 \\
    CLOM \cite{CLOM} & 79.57 & 76.07 & 72.94 & 69.82 & 67.80 & 65.56 & 63.94 & 62.59 & 60.62 & 60.34 & 59.58 & 67.16 \\
    LIMIT \cite{FSCILMultiPhaseTasks} & 75.89 & 73.55 & 71.99 & 68.14 & 67.42 & 63.61 & 62.40 & 61.35 & 59.91 & 58.66 & 57.41 & 65.48 \\
    MetaFSCIL \cite{MetaFSCILApproach} & 75.90 & 72.41 & 68.78 & 64.78 & 62.96 & 59.99 & 58.30 & 56.85 & 54.78 & 53.82 & 52.64 & 61.93 \\
    FACT \cite{FACT} & 75.90 & 73.23 & 70.84 & 66.13 & 65.56 & 62.15 & 61.74 & 59.83 & 58.41 & 57.89 & 56.94 & 64.42 \\
    PFR \cite{PanPFR} & 77.78 & 73.72 & 71.40 & 68.75 & 67.96 & 63.05 & 62.87 & 60.61 & 59.59 & 58.73 & 57.54 & 65.64 \\
    CLOSER \cite{CLOSER} & 79.40 & 75.92 & 73.50 & 70.47 & 69.24 & 67.22 & 66.73 & \underline{65.69} & \underline{64.00} & \underline{64.02} & \underline{63.58} & 69.07 \\
    
    SAVC \cite{SAVC} & $\mathbf{81.85}$ & 77.92 & 74.95 & 70.21 & 69.96 & 67.02 & 66.16 & 65.30 & 63.84 & 63.15 & 62.50 & 69.35 \\
    NC-FSCIL \cite{NeuralCollapseFSCIL} & 80.45 & 75.98 & 72.30 & 70.28 & 68.17 & 65.16 & 64.43 & 63.25 & 60.66 & 60.01 & 59.44 & 67.28 \\
    GKEAL \cite{GKEALFramework} & 78.88 & 75.62 & 72.32 & 68.62 & 67.23 & 64.26 & 62.98 & 61.89 & 60.20 & 59.21 & 58.67 & 66.35 \\
    CABD \cite{ClassAwareDistillation} & 79.12 & 75.37 & 72.80 & 69.05 & 67.53 & 65.12 & 64.00 & 63.51 & 61.87 & 61.47 & 60.93 & 67.33 \\ 
    OrCo \cite{OrCo} & 75.59 & 66.84 & 65.02 & 63.01 & 61.57 & 59.41 & 58.76 & 58.36 & 56.76 & 57.66 & 57.15 & 62.36 \\ 
    Comp-FSCIL \cite{CompositionalFSCIL} & 80.94 & 77.51 & 74.34 & 71.00 & 68.77 & 66.41 & 64.85 & 63.92 & 62.12 & 62.10 & 61.17 & 68.47 \\ \hline
    $\mathbf{MoTiC\ (Ours)}$ & \underline{81.35} & $\mathbf{78.17}$ & $\mathbf{75.80}$ & $\mathbf{72.91}$ & $\mathbf{71.52}$ & $\mathbf{69.49}$ & $\mathbf{68.60}$ & $\mathbf{67.36}$ & $\mathbf{65.54}$ & $\mathbf{65.63}$ & $\mathbf{65.27}$ & $\mathbf{70.79}$ \\ \hline
    \end{tabular}%
    }
    
\end{table*}

\section{Experiments}
\label{Experiments}

\begin{table*}[t]
    \caption{5-way 5-shot performance on miniImageNet using 
    the Resnet18 backbone}
    \label{tab:results_miniImageNet}
    \centering
    
    \resizebox{\textwidth}{!}{
    \begin{tabular}{@{}lcccccccccc@{}}
    \toprule
    \multirow{2}{*}{Method} & \multicolumn{9}{c}{Acc. in each session (\%)} & \multirow{2}{*}{Avg (\%) } \\ \cmidrule(r){2-10}
     & 0 & 1 & 2 & 3 & 4 & 5 & 6 & 7 & 8 &  \\ \midrule
    Baseline & 72.27 & 67.46 & 63.26 & 59.73 & 56.56 & 53.53 & 50.90 & 48.93 & 47.26 & 57.77 \\ \midrule
    TOPIC \cite{FewShotCIL} & 61.31 & 50.09 & 45.17 & 41.16 & 37.48 & 35.52 & 32.19 & 29.46 & 24.42 & 39.64 \\
    F2M \cite{FlatMinima} & 67.28 & 63.80 & 60.38 & 57.06 & 54.08 & 51.39 & 48.82 & 46.58 & 44.65 & 54.89 \\
    CEC \cite{EvolvedClassifiers} & 72.00 & 66.83 & 62.97 & 59.43 & 56.70 & 53.73 & 51.19 & 49.24 & 47.63 & 57.74 \\
    ALICE \cite{OpenSetFSCIL} & 80.60 & 70.60 & 67.40 & 64.50 & 62.50 & 60.00 & 57.80 & 56.80 & 55.70 & 63.98 \\
    CLOM \cite{CLOM} & 73.08 & 68.09 & 64.16 & 60.41 & 57.41 & 54.29 & 51.54 & 49.37 & 48.00 & 58.48 \\
    LIMIT \cite{FSCILMultiPhaseTasks} & 72.32 & 68.47 & 64.30 & 60.78 & 57.95 & 55.07 & 52.70 & 50.72 & 49.14 & 59.05 \\
    MetaFSCIL \cite{MetaFSCILApproach} & 72.04 & 67.94 & 63.77 & 60.29 & 57.58 & 55.16 & 52.90 & 50.79 & 49.19 & 58.85 \\
    FACT \cite{FACT} & 72.56 & 69.63 & 66.38 & 62.77 & 60.60 & 57.33 & 54.34 & 52.16 & 50.49 & 60.70 \\
    GKEAL \cite{GKEALFramework} & 73.59 & 68.90 & 65.33 & 62.29 & 59.39 & 56.70 & 54.20 & 52.59 & 51.31 & 60.48 \\
    CABD \cite{ClassAwareDistillation} & 74.65 & 70.43 & 66.29 & 62.77 & 60.75 & 57.24 & 54.79 & 53.65 & 52.22 & 61.42 \\ 
    CLOSER \cite{CLOSER} & 76.02 & 71.61 & 67.99 & 64.69 & 61.70 & 58.94 & 56.23 & 54.52 & 53.33 & 62.78 \\ 
    OrCo \cite{OrCo} & $\mathbf{83.30}$ & 75.32 & 71.53 & 68.16 & 65.63 & 63.12 & \underline{60.20} & \underline{58.81} & $\mathbf{58.08}$ &  \underline{67.14} \\ 
    SAVC \cite{SAVC} & 81.12 & \underline{76.14} & \underline{72.43} & \underline{68.92} & \underline{66.48} & \underline{62.95} & 59.92 & 58.39 & 57.11 & 67.05 \\ \midrule
    $\mathbf{MoTiC (Ours)}$ & \underline{81.27} & $\mathbf{76.63}$ & $\mathbf{72.78}$ & $\mathbf{69.69}$ & $\mathbf{66.94}$ & $\mathbf{63.86}$ & $\mathbf{61.1}$ & $\mathbf{59.15}$ & \underline{57.70} & $\mathbf{67.68}$ \\ \bottomrule
    \end{tabular}%
    }
    \end{table*}

\begin{figure}[t]
    \centering
    \includegraphics[width=1.0\textwidth]{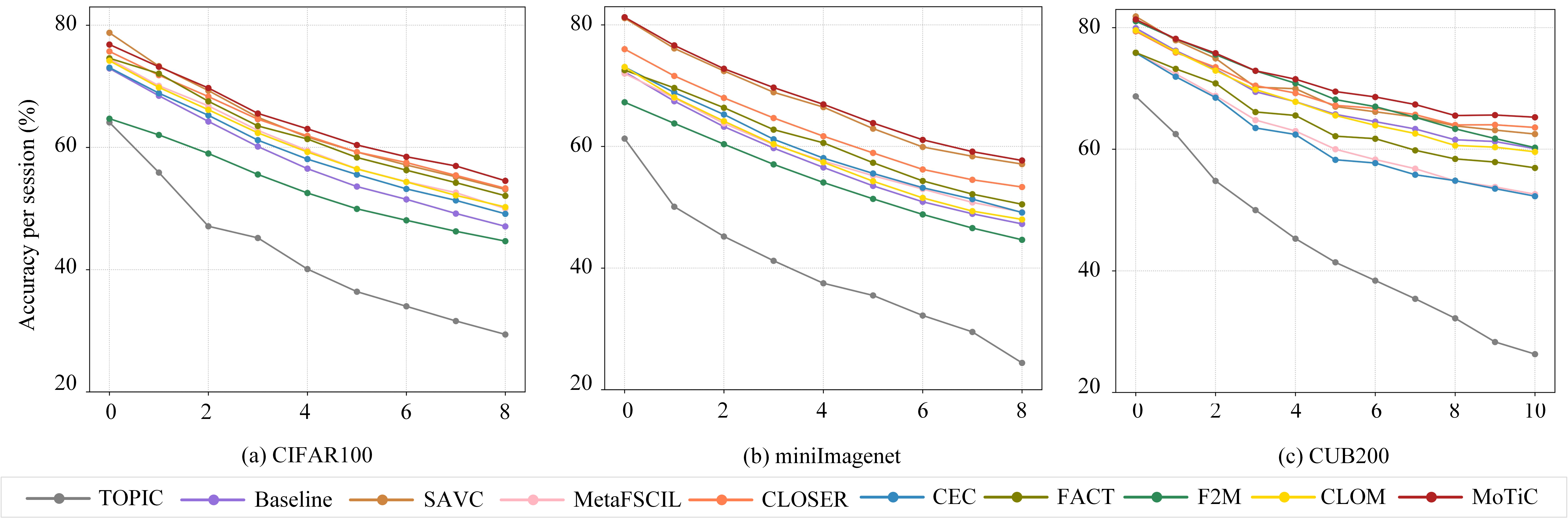}
    \caption{SOTA comparisons on the CIFAR100, miniImageNet, and CUB200 benchmarks.}\label{fig1}
    \end{figure}

\subsection{Experimental setting}
\label{Experimental setting}

\subsubsection{Datasets}
\label{Datasets}
We evaluate MoTiC on the CUB200, CIFAR100, and miniImageNet \cite{CUB2011,CIFAR100,miniImageNet} datasets. The classes in each dataset are divided into base classes with sufficient annotated samples and new classes with $K$ labelled images. For FSCIL, the network is trained on base classes during the first session, whereas few-shot new classes are gradually introduced in $T$ subsequent sessions.

In each incremental session, $N$-way $K$-shot new classes are added. CIFAR100 and miniImageNet consist of 100 classes, with 60 as base classes and 40 as new classes. Each base class contains 500 training samples, whereas each new class contains 5 training samples ($K$=5), with 100 test samples per class. CUB200 consists of 200 classes, where 100 classes are designated as base classes and another 100 as new classes, with $K$ =5, $T$ = 10, and $N$ =10.

\subsubsection{Implementation}
\label{Implementation}
Our method is conducted with the PyTorch library with an SGD
optimizer with Nesterov momentum 0.9.
Following the method of Zhang et al. \cite{EvolvedClassifiers}, we use ResNet-18 \cite{DeepResidualLearning} for the miniImageNet and CUB200 experiments and ResNet-20 \cite{DeepResidualLearning} for the CIFAR100 experiments. We adhere to the convention of using a ResNet-18 model pretrained on the ImageNet dataset for CUB200 experiments. We set the mini-batch size to 128, 128, and 64 for the CIFAR100, miniImageNet, and CUB200 experiments, respectively. We configure the hyperparameters as follows: for CIFAR100, miniImageNet, and CUB200, we set $\lambda_{ssc}$ as 0.1, 0.1, and 0.01; $\lambda_{MoTi}$ as 2.5, 1.5, and 2.5; and the queue sizes as 65536, 8192, and 8192, respectively.
Additionally, we configure different rotation angle combinations for the virtual classes: for CIFAR100 and CUB200, the angles are set to 0° and 180°; for miniImageNet, the angles are set to 0°, 90°, 180°, and 270°.
Cosine scheduling is employed with a maximum learning rate set to 0.1. These hyperparameters are determined through
a simple grid search.

\begin{table*}[!]
    \caption{5-way 5-shot performance on CIFAR100 using 
    the Resnet20 backbone}
    \label{tab:results_CIFAR100}
    \centering
    
    \resizebox{\textwidth}{!}{
    \begin{tabular}{@{}lcccccccccc@{}}
    \toprule
    \multirow{2}{*}{Method} & \multicolumn{9}{c}{Acc. in each session (\%)} & \multirow{2}{*}{Avg (\%) } \\ \cmidrule(r){2-10}
     & 0 & 1 & 2 & 3 & 4 & 5 & 6 & 7 & 8 &  \\ \midrule
    Baseline & 72.93 & 68.46 & 64.26 & 60.15 & 56.53 & 53.60 & 51.51 & 49.19 & 47.09 & 58.19 \\ \midrule
    TOPIC \cite{FewShotCIL} & 64.10 & 55.88 & 47.07 & 45.16 & 40.11 & 36.38 & 33.96 & 31.55 & 29.37 & 42.62 \\
    CEC \cite{EvolvedClassifiers} & 73.07 & 66.88 & 65.26 & 61.19 & 58.09 & 55.57 & 53.22 & 51.34 & 49.14 & 59.53 \\

    CLOM \cite{CLOM} & 74.20 & 69.83 & 66.17 & 62.39 & 59.26 & 56.48 & 54.36 & 52.16 & 50.25 & 60.57 \\
    
    LIMIT \cite{FSCILMultiPhaseTasks} & 73.81 & 72.09 & 67.87 & 63.89 & 60.70 & 57.77 & 55.67 & 53.52 & 51.23 & 61.83 \\
    MetaFSCIL \cite{MetaFSCILApproach} & 74.50 & 70.10 & 66.84 & 62.77 & 59.48 & 56.52 & 54.36 & 52.56 & 49.97 & 60.79 \\

    FACT \cite{FACT} & 74.60 & 72.09 & 67.56 & 63.52 & 61.38 & 58.36 & 56.28 & 54.24 & 52.10 & 62.23 \\

    GKEAL \cite{GKEALFramework} & 74.01 & 70.45 & 67.01 & 63.08 & 60.01 & 57.30 & 55.50 & 53.39 & 51.40 & 61.35 \\
    CLOSER \cite{CLOSER} & 76.02 & 71.61 & 67.99 & 64.69 & 61.70 & 58.94 & 56.23 & 54.52 & \underline{53.33} & 63.10 \\ 
    OrCo \cite{OrCo} & $\mathbf{80.08}$ & 68.16 & 66.99 & 60.98 & 59.78 & 58.60 & 57.04  &  55.13 & 52.19 & 62.11 \\
    SAVC \cite{SAVC} & \underline{78.77} & $\mathbf{73.31}$ & 69.31 & 64.93 & 61.70 & 59.25 & 57.13 & 55.19 & 53.12 & \underline{63.63} \\ \midrule
    $\mathbf{MoTiC (Ours)}$ & 76.85 & \underline{73.2} & $\mathbf{69.73}$ & $\mathbf{65.57}$ & $\mathbf{63.05}$ & $\mathbf{60.42}$ & $\mathbf{58.48}$ & $\mathbf{56.97}$ & $\mathbf{54.55}$ & $\mathbf{64.31}$ \\ \bottomrule
    \end{tabular}%
    }
    
    \end{table*}
    
\subsubsection{Evaluation}
\label{Evaluation}
We use the accuracy of the last session as the evaluation metric, including base class accuracy ($A_B$), new class accuracy ($A_N$), and overall accuracy ($A_W$). These metrics assess the following in the ablation study: feature distinguishability, transferability, and their trade-offs. Additionally, a global accuracy ($A_{avg}$) is calculated by averaging accuracies across all sessions to compare the average performance with prior studies. All the MoTiC results are derived from averaging three trials.

\subsection{Comparison with state-of-the-art methods}
\label{Comparison with state-of-the-art methods}
We compare our proposed MoTiC method with prior methods on the CUB200 (Table \ref{tab:cub200_results}), miniImageNet (Table \ref{tab:results_miniImageNet}), and CIFAR100 (Table \ref{tab:results_CIFAR100}) datasets. The results indicate that MoTiC achieves state-of-the-art performance across all three datasets, with notable superiority on the fine-grained CUB200 dataset. Specifically, it exceeds the second-best method by 1.31\% in $A_{avg}$ and by 1.69\% in $A_{W}$, as shown in Table \ref{tab:cub200_results}. This significant improvement is because, in fine-grained datasets, the prior information from old classes is more abundant and useful for new classes than those in coarse-grained datasets.


\subsection{Ablation studies}
\label{Ablation Studies}
In our ablation study presented in Table 4, we evaluate three key components to validate the importance of our proposed modules. Within the incremental-frozen framework, we establish a baseline using only the cross-entropy (CE) loss and cosine similarity metric. We then investigate the impact of three critical components: the momentum contrastive self-supervision module, the momentum tightness module, and the virtual class module. Additionally, we examine the influence of the projection head, a common element in contrastive learning. Our results indicate that the optimal configuration is achieved by combining 
$\mathcal{L}_{ssc(MoCo)}$, $\mathcal{L}_{MoTi}$, and the virtual class module, which yields the highest overall performance with an $A_{Avg}$ of 64.31\%. This combination effectively leverages the foundational enhancements provided by $\mathcal{L}_{ssc(MoCo)}$ and $\mathcal{L}_{MoTi}$
, further refines the performance through the virtual class module, and avoids potential negative impacts associated with the projection head.

\subsection{Quantitative Analysis}
\label{Quantitative Analysis}
\begin{figure}[t]
    \centering
    \includegraphics[width=1.0\textwidth]{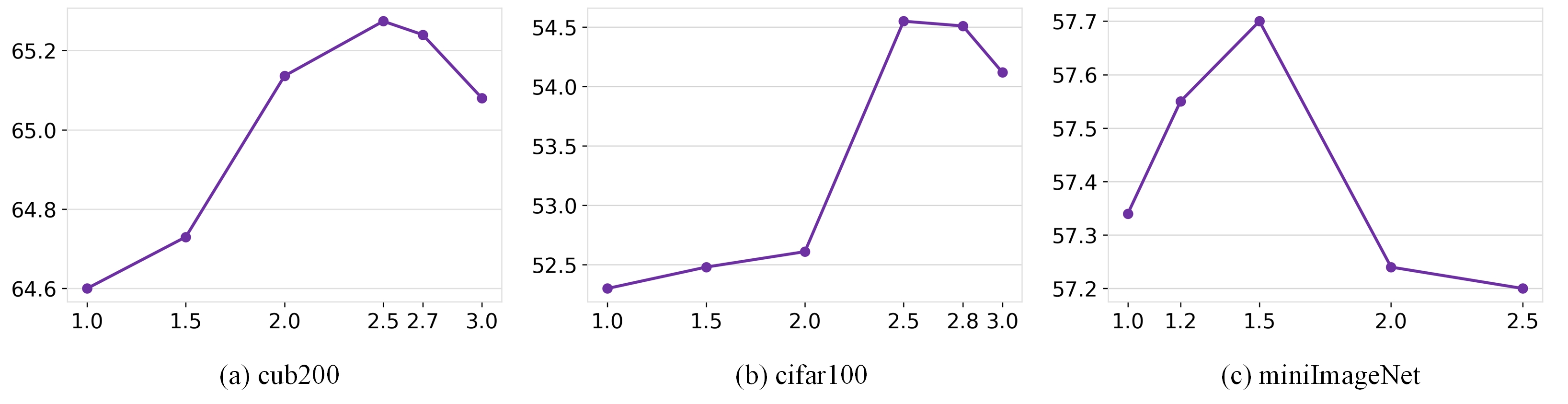}
    \caption{Sensitivity study of the momentum tightness hyperparameter $\lambda_{MoTi}$,
    illustrating the model's $A_{W}$ across varying values of $\lambda_{MoTi}$. }
    \label{fighyperparameter}
\end{figure}

In the quantitative analysis, we utilize $\mathcal{L}_{MoTi}$ to regulate the interclass similarity among virtual classes, and we report 
the relationship between performance and interclass distances in Fig. \ref{fighyperparameter}. It can be observed that all three datasets exhibit 
a similar trend in controlling interclass similarity, indicating the demand to adjust $\mathcal{L}_{MoTi}$ according to different tasks.

\definecolor{darkred}{RGB}{178, 34, 34} 

\begin{table}[t]
    \caption{Ablation studies on CIFAR100.}
    \label{tab:ablation_cifar100}
    \centering
    
    \resizebox{\textwidth}{!}{
    \begin{tabular}{@{}cccccccc@{}}
    \toprule
    $\mathcal{L}_{ssc(MoCo)}$ & $\mathcal{L}_{MoTi}$ & $projector$ & $virtual\ class$  & $A_B$ (\%) & $A_N$ (\%) & $A_W$ (\%) & $Avg$ (\%) \\ \midrule
    \ding{55}       & \ding{55}        & \ding{55}     &\ding{55}          & 68.02      & 20.95      & 50.46      & 24.81   \\
    \textcolor{darkred}{\checkmark}        & \ding{55}  &\ding{55}      & \ding{55}        & 67.68      & 26.5      & 51.21      & 61.11   \\
    \ding{55}        & \textcolor{darkred}{\checkmark}  &\ding{55}       & \ding{55}       & 70.53      & 22.78      & 51.43      & 62.48   \\

    \textcolor{darkred}{\checkmark}        & \textcolor{darkred}{\checkmark}        & \textcolor{darkred}{\checkmark} &\ding{55}       & 67.5      & 25.45      & 50.68      & 60.91   \\
    \textcolor{darkred}{\checkmark}       & \textcolor{darkred}{\checkmark}          & \ding{55}  &\ding{55}       & 71.52      & 25.48      & 53.1      & 63.53   \\
    
    \textcolor{darkred}{\checkmark}      & \ding{55}  & \ding{55} & \textcolor{darkred}{\checkmark} &  70.28      & 28.72      & 53.66      & 62.99   \\

    \ding{55}      & \textcolor{darkred}{\checkmark}  & \ding{55} & \textcolor{darkred}{\checkmark} &  72.3      & 26.35      & 53.92      & 64.02   \\
    
    \textcolor{darkred}{\checkmark}       & \textcolor{darkred}{\checkmark}       & \ding{55}  &\textcolor{darkred}{\checkmark}      & 70.72      & 27.23      & \textbf{54.55}      & \textbf{64.31}   \\ \bottomrule
    \end{tabular}
    }
\end{table}

\begin{figure}[t]
    \centering
    \includegraphics[width=1.0\textwidth]{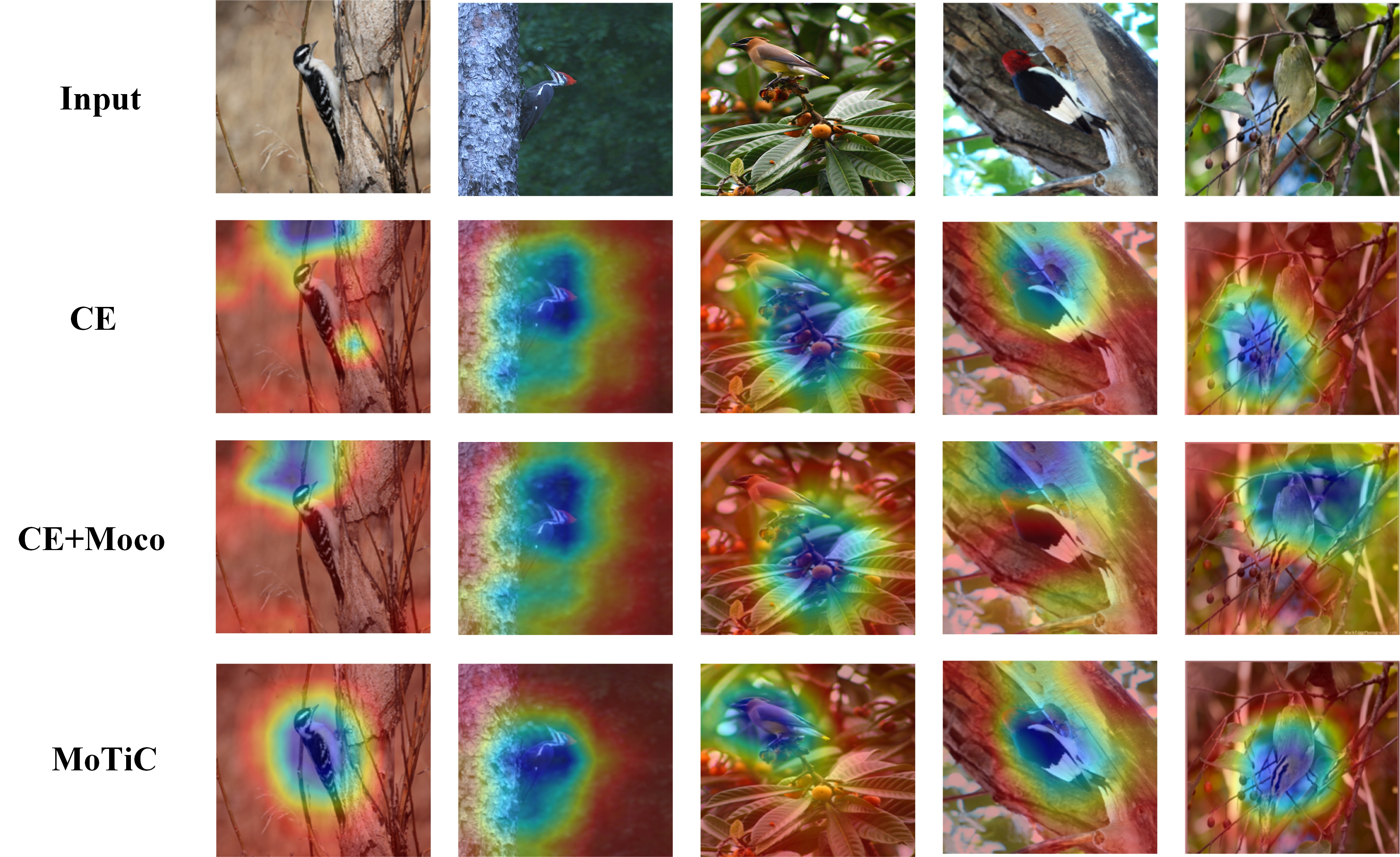}
    \caption{Visualization of the class-activation-map (CAM). CE: Baseline model; CE+MoCo: The combination of baseline and momentum self-supervised methods; MoTiC: Our proposed model. }
    \label{cam}
\end{figure}

\subsection{Visualization}
\label{Visualization}
To qualitatively assess the efficacy of our approach, we compare the baseline model and our MoTiC framework
using class activation maps (CAMs) (Zhou et al., 2016) \cite{DeepFeaturesDiscriminativeLocalization} and t-SNE \cite{TSNEVisualization} analysis.
Based on the CAM visualizations presented in Fig. \ref{cam},
while CE+Moco enhances the model's transfer learning performance, its prototypes under few-shot sampling fail to precisely represent the entire category. In contrast, MoTiC-CAM activates smaller regions with sharper focus on target content and better filtering of irrelevant backgrounds. 
This phenomenon substantiates the ability of the MoTiC framework to effectively increase the accuracy of few-shot sampling prototypes of novel classes, thereby enhancing the performance of FSCIL.

\begin{figure}[t]
    \centering
    \includegraphics[width=1.0\textwidth]{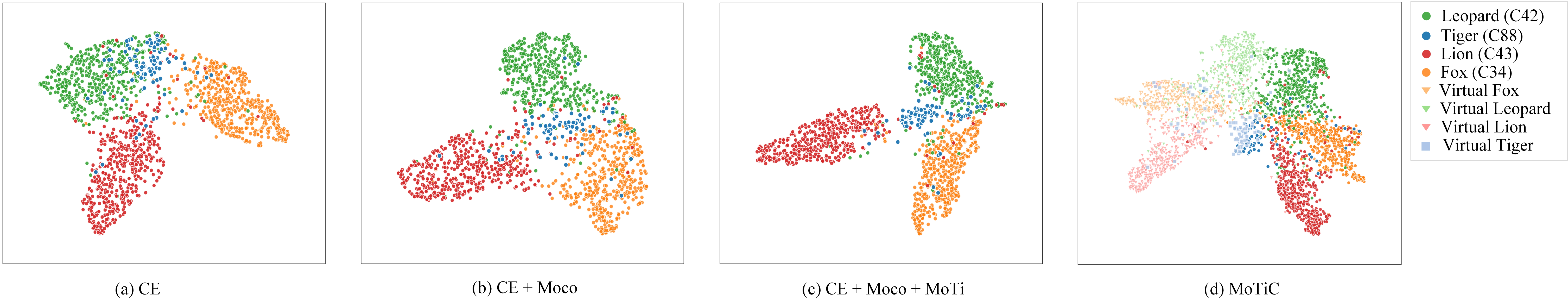}
    \caption{T-SNE visualization of learned representations on CIFAR100.  }
    \label{t-sne}
\end{figure}

To further analyse the effects of our proposed method, we visualize the learned representations trained with different configurations of the loss function, as illustrated in Fig. \ref{t-sne}.
We select a single new class, Tiger, and the three base classes with the most similar feature space characteristics:
Leopard, Fox, and Lion. In the baseline representation (a), Tiger significantly overlaps with the existing class Leopard.
Momentum self-supervised learning (b) enhances separation between the new and old classes
but increases intraclass variability, reducing prototype precision in few-shot sampling and
lowering base-class discriminability. The momentum tightness method (c) improves compactness for both the new and base classes, 
boosting their distinctiveness.
The virtual class method (d), which is compatible with previous approaches, further refines class separation.

\begin{figure}[t]
    \centering
    \includegraphics[width=1.0\textwidth]{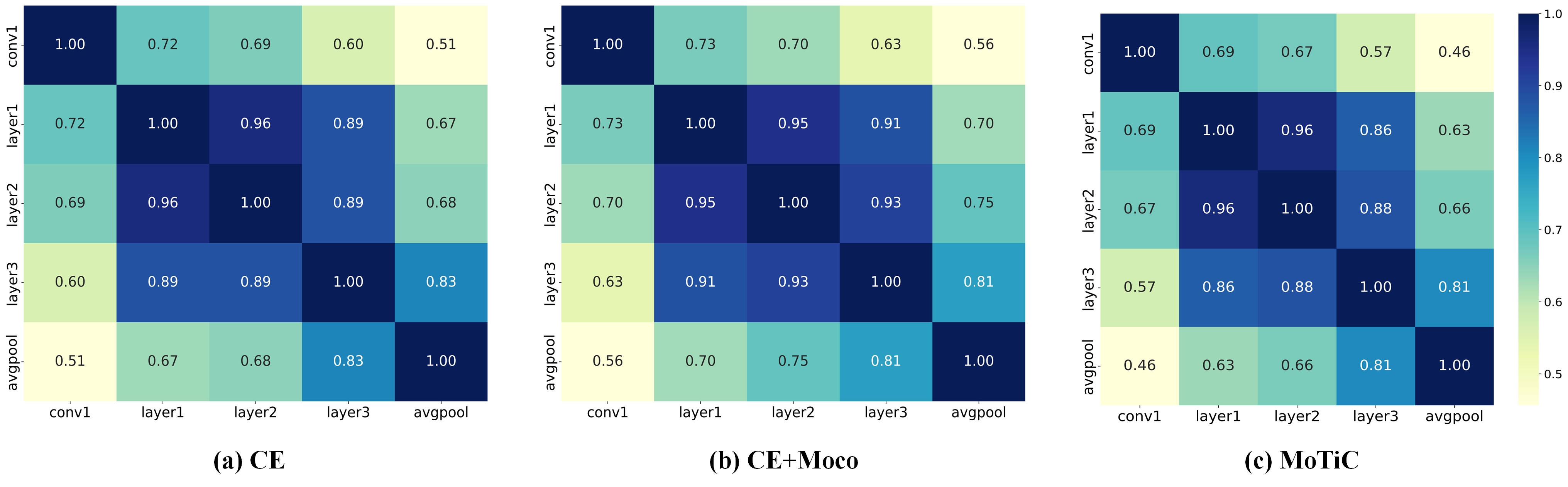}
    \caption{Centred Kernel Alignment (CKA) scores.  }
    \label{CKA}
\end{figure}

Finally, we compared the CKA \cite{SimilarityNNRepresentations} (centred kernel alignment) similarity across
different network layers to evaluate the variations induced by different loss combinations.
As illustrated in Fig. \ref{CKA}, the application of momentum contrastive loss significantly increases CKA similarity, indicating that the backbone network captures patterns more analogous to simpler visual primitives such as edges or corners.
This observation validates that the model tends to learn elementary patterns that are more readily shared across categories.
However, when momentum tightness loss is integrated, the interlayer CKA similarity noticeably decreases;
this demonstrates that the model captures more sophisticated and discriminative feature representations while maintaining
the generalization capability inherited from self-supervised learning. The combined strategy achieves
enhanced feature distinctiveness through complex pattern encoding,
thereby striking an optimal balance between task-specific discriminative power
and cross-domain representational versatility.

\section{Conclusion}
\label{Conclusion}
To increase the accuracy of few-shot sampling prototypes in FSCIL, we present a novel contrastive learning methodology
that modulates the richness and compactness of features across various classes within the encoder and
synergizes with virtual class representations. Extensive experiments and in-depth analyses demonstrate that our approach establishes a new, significantly improved state-of-the-art.
In the future, we will explore more effective strategies to synthesize novel classes by combining features from existing categories.
Additionally, while we have evaluated our framework exclusively in the context of FSCIL,
we believe that its principles, particularly those promoting generalization in representation learning, could benefit other domains
that require robust and transferable representations.

\end{document}